\documentclass[letterpaper]{article} 
\usepackage{aaai25}  
\usepackage{times}  
\usepackage{helvet}  
\usepackage{courier}  
\usepackage[hyphens]{url}  
\usepackage{graphicx} 
\usepackage{natbib}  
\usepackage{caption} 
\usepackage{algorithm}
\usepackage{algorithmic}
\usepackage{amsfonts}
\usepackage{newfloat}
\usepackage{listings}
\usepackage{multirow}
\usepackage{booktabs}
\usepackage{pifont}
\usepackage{amsmath}
\usepackage{newfloat}
\usepackage{listings}
\DeclareCaptionStyle{ruled}{labelfont=normalfont,labelsep=colon,strut=off} 
\floatstyle{ruled}
\newfloat{listing}{tb}{lst}{}
\floatname{listing}{Listing}
\title{MegActor-$\Sigma$: Unlocking Flexible  Mixed-Modal Control in Portrait Animation with Diffusion Transformer}

\author {
    Shurong Yang\footnotemark[1]$^{,1}$\ ,
    Huadong Li\footnotemark[1]$^{,1}$\ ,
    Juhao Wu\footnotemark[1]$^{,1}$\ ,
    Minhao Jing\footnotemark[1]$^{,1}$
    Linze Li$^{,1}$,  \\
    Renhe Ji\footnotemark[2]$^{,1}$,
    Jiajun Liang\footnotemark[2]$^{,1}$,
    Haoqiang Fan$^{1}$, 
    Jin Wang$^{2}$, \\
}
\affiliations{
    \textsuperscript{\rm 1}MEGVII Technology,
    \textsuperscript{\rm 2}The University of Hong Kong\\

    \tt\small $\{$lihuadong, wujuhao, lilinze, jirenhe, liangjiajun, fanhaoqiang$\}$@megvii.com \\
    \tt\small $\{$yangshurong6894, jingminhao666, wjbillbieber$\}$@gmail.com \\ 
}

\usepackage{bibentry}

\begin{document}



\twocolumn[{%
\renewcommand\twocolumn[1][]{#1}%
\maketitle
\vspace{-8mm}
\begin{center}
    \centering
    \includegraphics[width=0.95\linewidth]{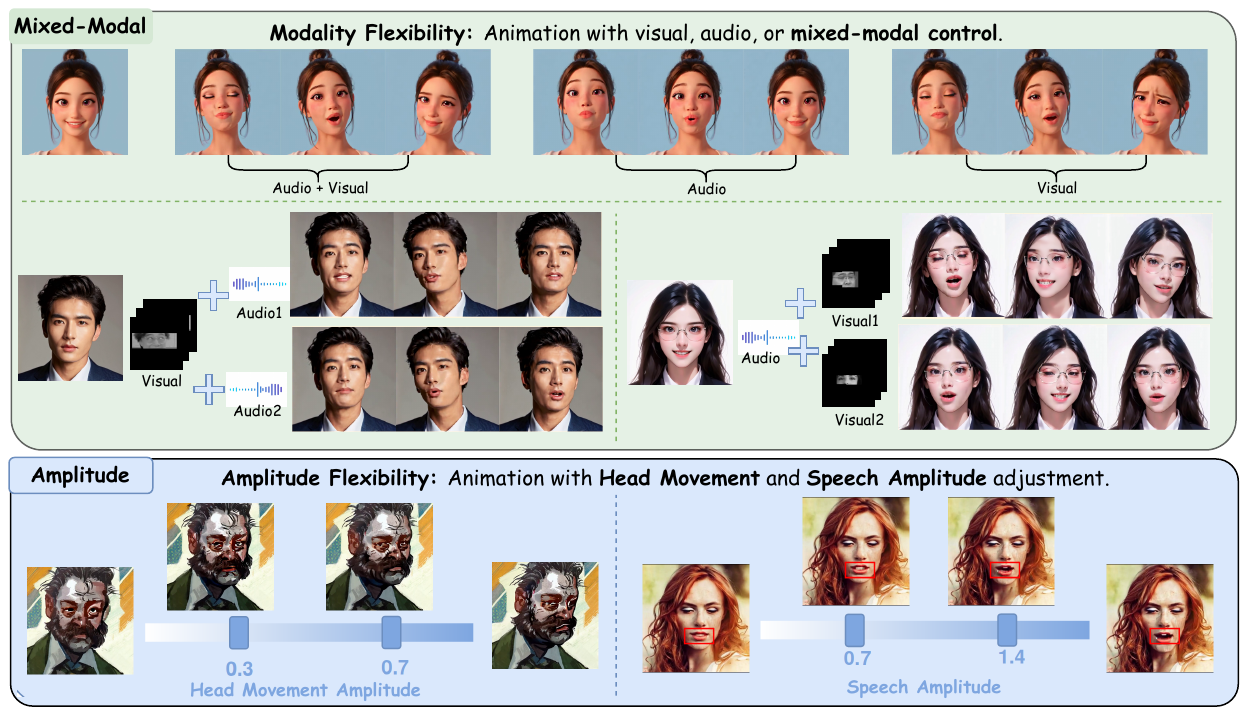}
    \captionof{figure}{
    Qualitative results of MegActor-$\Sigma$ in generating high-quality and flexible portrait animations, include: 1) \textit{Modality Flexibility}, enabling control through visual, audio or mixed-modal control;
    2) \textit{Amplitude Flexibility}, enabling adjustment of the scale of head movement and speech amplitude. Moreover, MegActor-$\Sigma$ is trained purely on public datasets, which successfully outperforms previous closed-source methods. Please see our project page for detailed comparisons.
    }
\label{fig:visual}
\end{center}%
}]

{
  \renewcommand{\thefootnote}{\fnsymbol{footnote}}
  \footnotetext[1]{Indicates equal contribution.}
  \footnotetext[2]{Corresponding author.}
}

\begin{abstract}
Diffusion models have demonstrated superior performance in the field of portrait animation. 
However, current approaches relied on either visual or audio modality to control character movements, failing to exploit the potential of mixed-modal control.
This challenge arises from the difficulty in balancing the weak control strength of audio modality and the strong control strength of visual modality.
To address this issue, we introduce MegActor-$\Sigma$: a mixed-modal conditional diffusion transformer (DiT), which can flexibly inject audio and visual modality control signals into portrait animation.
Specifically, we make substantial advancements over its predecessor, MegActor, by leveraging the promising model structure of DiT and integrating audio and visual conditions through advanced modules within the DiT framework.
To further achieve flexible combinations of mixed-modal control signals, we propose a ``Modality Decoupling Control" training strategy to balance the control strength between visual and audio modalities, 
along with the ``Amplitude Adjustment" inference strategy to freely regulate the motion amplitude of each modality.
Finally, to facilitate extensive studies in this field, we design several dataset evaluation metrics to filter out public datasets and solely use this filtered dataset to train MegActor-$\Sigma$.
Extensive experiments demonstrate the superiority of our approach in generating vivid portrait animations, outperforming previous methods trained on private dataset.
\begin{links}
    \link{Project Page}{https://megactor-ops.github.io/}
\end{links}
\end{abstract}

\section{Introduction}
Portrait animation refers to the task of animating a static portrait image using motions and facial expressions from a driving video, while preserving the identity and background of the portrait image.
This field has garnered significant attention due to its wide range of applications, including digital avatars~\cite{ma2021pixel, wang2021one} and AI-based human conversations~\cite{aicommunicate, johnson2018assessing}.
In recent years, diffusion models~\cite{xie2024x,tian2024emo} have showcased their advantages in portrait animation domain with single-modality control, whether through audio or visual modality.

Previous methods for generating portrait animations with visual modality control~\cite{xu2023magicanimate, zhu2024champ,yang2024megactor} typically resorted to intermediate motion representations extracted from the driving video, such as landmarks, dense poses, or face meshes. 
Though demonstrating satisfactory generation quality, obtaining such detailed visual modality control signals in real life scenarios can be inflexible and may require considerable human efforts.
Meanwhile, audio-driven methods \cite{wei2024aniportrait, wang2024instructavatar, musetalk} allowed users to control portrait animations using audio, supplemented by additional signals like blinking and head rotation.
However, these control signals were often insufficient for accurately capturing facial expressions, such as subtle movements of the eyebrows and lips, limiting the model's performance.

\begin{figure}
    \centering
    \includegraphics[width=0.99\linewidth]{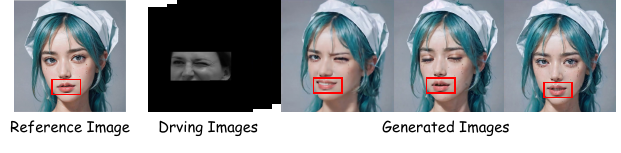}
    \caption{
    The visualization of visual leakage. Even when we remove mouth-driven components in visual modality, as V-Express~\cite{wang2024v} does, the generated results still exhibit a certain pattern of speaking without audio-driven. 
    }
    \label{fig:unbalance}
\end{figure}

To achieve flexible control with superior performance simultaneously, it is essential to endow the model with the ability to freely combine visual and audio modalities as control signals. 
To this end, the audio modality provides low-cost and flexible lip control, while the visual modality offers accurate motion and facial expression. 
However, training a model with such comprehensive capabilities poses significant challenges.
Previous methods \cite{wang2024v} have shown that the audio modality often struggles to exert control due to the dominance of the visual modality control.
As illustrated in Fig. \ref{fig:unbalance}, we observe that even when only partial visual control signals are provided (\emph{i.e.}, eye patches), the mouth regions still display a speaking pattern \cite{yang2024megactor,xie2024x}, indicating strong coherence of different facial regions causes visual leakage and prevents the mouth area from following the audio signal cues. Thus, learning a combination of two modalities requires not only the decoupling of modalities, but also the proactive spatial decoupling of visual modality. 

To fully explore the potential of decoupled modality control, we present MegActor-$\Sigma$, a mixed-modal conditional diffusion transformer (DiT) built upon its predecessor, MegActor \cite{yang2024megactor}.
Compared to SD1.5, the DiT architecture has fewer parameters, stronger pretrained weights, and produces more stable portrait animations.
We further introduce a 3-stage ``Modality Decoupling Control" training strategy to prevent visual modality leakage and address the imbalanced control strength of mixed modalities.
The first stage, ``Spatial-Decoupling Visual Training", aims to spatially decouple visual control. We adopt face dropout strategy to generate spatial patch masks that isolate visual signals, such as those from the eyes or mouth. This method ensures that when eye signals are input, there is no control over the mouth area, thus preventing visual modality leakage.
The second stage, ``Modality-Decoupling Mixed Training", incorporates audio control while preventing it from being overwhelmed by the visual modality. This involves training with visual, audio and audio-visual mixed signals, while applying face dropout on the visual modality to balance its strength .
The final stage, ``Motion Priors Training", introduces temporal modules to enable temporal reasoning and consistency. It trains the temporal modules on audio, visual, and mixed modalities without face dropout on the visual modality.

Besides, we introduce a simple yet effective ``Amplitude Adjustment" inference strategy to arbitrarily regulate the motion amplitude. 
For the visual modality, warp transform mapping is used to project the driving images towards a central position, adjusting the magnitude of motion. 
For the audio modality, the control strength can be adjusted by modifying the combination ratio after the audio attention mechanism, thus enabling customizable speaking amplitude.

Finally, to facilitate the development of the open-source community in this field, we design several dataset evaluation metrics to filter out public datasets.
MegActor-$\Sigma$ is trained solely on this filtered public dataset, which consists of 100 hours of content featuring high-resolution facial regions, high lip-sync accuracy, and a high proportion of frontal facial orientations. 
Extensive experiments demonstrate the superior performance of our approach in generating vivid portrait animations, outperforming previous closed-source methods.
Our contributions are summarized as follows:

\begin{itemize}
    \item A novel mixed-modal Diffusion Transformer (DiT) that effectively integrates audio and visual control signals. To the best of our knowledge, this is the first portrait animation method based on the framework of DiT, compared to previous UNet-based methods.
    \item A novel ``Modality Decoupling Control" training strategy that solves visual leakage and effectively balances the control strength between visual and audio modality.
    \item A set of quality evaluation metrics for filtering public multimodal portrait animation datasets and a filtered 100-hour high-quality dataset for open-source research.
    \item Extensive experiments demonstrate that our approach excels in generating vivid portrait animations and offers superior flexibility in its application.
\end{itemize}

\section{Related Work}
\noindent \textbf{GAN-based Portrait Animation.} Traditional portrait animation methods often employ neural networks to decouple and extract motion features from audio or visual modalities, converting these features into intermediate representations such as landmarks~\cite{su2024audio, song2022audio}, 3D head parameters (e.g., 3d Face Morphable Model~\cite{zhang2024learning, ma2023styletalk, xing2023codetalker, bai2024efficient} (3DMM~\cite{tran2018nonlinear}), Faces Learned with an Articulated Model and Expressions~\cite{ma2024cvthead} (FLAME~\cite{li2017learning}), 3D Gaussian parameters~\cite{cho2024gaussiantalker, chen2024gstalker, li2024talkinggaussian, zhuang2024learn2talk}, 3D Tri-Plane Hash Representation~\cite{li2023efficient}), or latent representations like motion keypoints~\cite{song2024adaptive, gan2023efficient, tan2023emmn, xu2023high, ji2022eamm} and feature encodings~\cite{drobyshev2024emoportraits, bi2024nerf, peng2024synctalk, li2024ae, su2024dt, liu2023moda, guan2023stylesync, wang2023seeing, liang2022expressive}. Techniques such as GANs~\cite{karras2019style}, NeRF~\cite{mildenhall2021nerf}, and Gaussian Splatting~\cite{kerbl20233d} are then used to render dynamic portrait animations.
Though initially successful, these works struggle to  capture accurate motion information and generate precise and realistic control signals for portrait animation.

\noindent \textbf{Diffusion-based Portrait Animation.} Recent advances have seen diffusion models excel in image and video generation~\cite{blattmann2023align, khachatryan2023text2videozero, luo2023videofusion, xu2024easyanimate, chen2023pixart, blattmann2023stable, guo2023animatediff}, prompting studies to explore their use in creating high-quality image or videos. 
Image-driven methods~\cite{hu2023animate, chang2023magicdance, wang2023disco} use facial landmarks and poses to extract motion for reconstruction but struggle with subtle expressions and can cause facial distortion when the driving signal and reference figure differ in identity and facial proportions, affecting expressiveness and stability.
Audio-driven Talking Head generation methods~\cite{xu2024hallo, stypulkowski2024diffused, xu2024vasa, liu2024anitalker, yu2024make} utilize audio for lip synchronization and weak visual signals for head motion, reducing facial distortion and producing natural videos. However, they fall short in controlling nuanced eye movements and may mismatch head motion amplitude to the driving video, and they often remain uni-modal, missing the benefits of multi-modal control.
V-Express~\cite{wang2024v} employs audio and weak image signals for multi-modal control but lacks flexibility due to the simultaneous requirement of both modalities.
EchoMimic~\cite{chen2024echomimic} generates portrait animations under single-modal or multi-modal control. However, it relies on Mediapipe~\cite{lugaresi2019mediapipe} for key points, leading to lost visual detail and high detector dependence. 
Additionally, EchoMimic follows the previous SD1.5 architecture, lacking to explore the potential of new frameworks.
In contrast, MegActor-$\Sigma$ utilizes the richest original driving images as the control signal and employs an advanced mixed-modal Diffusion Transformer, achieving outstanding performance in portrait animation. 
A summary of related work is presented in Table \ref{tab:related work}.

\begin{table}
    \centering
    \renewcommand{\arraystretch}{1.1}
    \resizebox{0.99\linewidth}{!}{
    \begin{tabular}{l c | ccc}
    \toprule[1.5pt]
    \multirow{2}[2]{*}{\textbf{Method}} & \multirow{2}[2]{*}{\textbf{Framework}} & \multicolumn{3}{c}{\textbf{Control Modality}} \\ \cmidrule[0.5pt](lr){3-5}
    & &  Audio & Visual & Audio + Visual \\
        \midrule[1pt]
        SadTalker~\cite{zhang2023sadtalker}  & \multirow{3}[2]{*}{GAN} & $\checkmark$ & $\times$ &  $\times$ \\
        LivePortarit~\cite{guo2024liveportrait} &  & $\times$ & $\checkmark$ &  $\times$ \\
        EMOPortarit~\cite{drobyshev2024emoportraits} &  & $\times$ & $\times$ &  $\checkmark$ \\
        \hline
        \begin{tabular}[c]{@{}l@{}}
        Hallo~\cite{xu2024hallo} \\
        EMO~\cite{tian2024emo} \\
        AniPortrait~\cite{wei2024aniportrait} \\
        \end{tabular} & Stable Diffusion (SD1.5)  &  $\checkmark$ & $\times$ & $\times$ \\
        \hline
        \begin{tabular}[c]{@{}l@{}}
        Follow-your-Emoji~\cite{ma2024follow} \\
        X-Portrait~\cite{xie2024x} \\
        MegActor~\cite{yang2024megactor} \\
        \end{tabular} & Stable Diffusion (SD1.5)  & $\times$ & $\checkmark$ & $\times$ \\
        \hline
        \begin{tabular}[c]{@{}l@{}}
        V-Express~\cite{wang2024v} \\
        \end{tabular} & Stable Diffusion (SD1.5)  & $\times$ & $\times$ & $\checkmark$ \\
        \midrule[1pt]
        \textbf{Ours} & Diffusion Transformer (DiT) & $\checkmark$ & $\checkmark$ & $\checkmark$ \\
    \bottomrule[1.5pt]
    \end{tabular}}
    \caption{Summary of the portrait animation methods.}
    \label{tab:related work}
\end{table}

\section{Methodology}
\begin{figure}
    \centering
    \includegraphics[width=0.98\columnwidth]{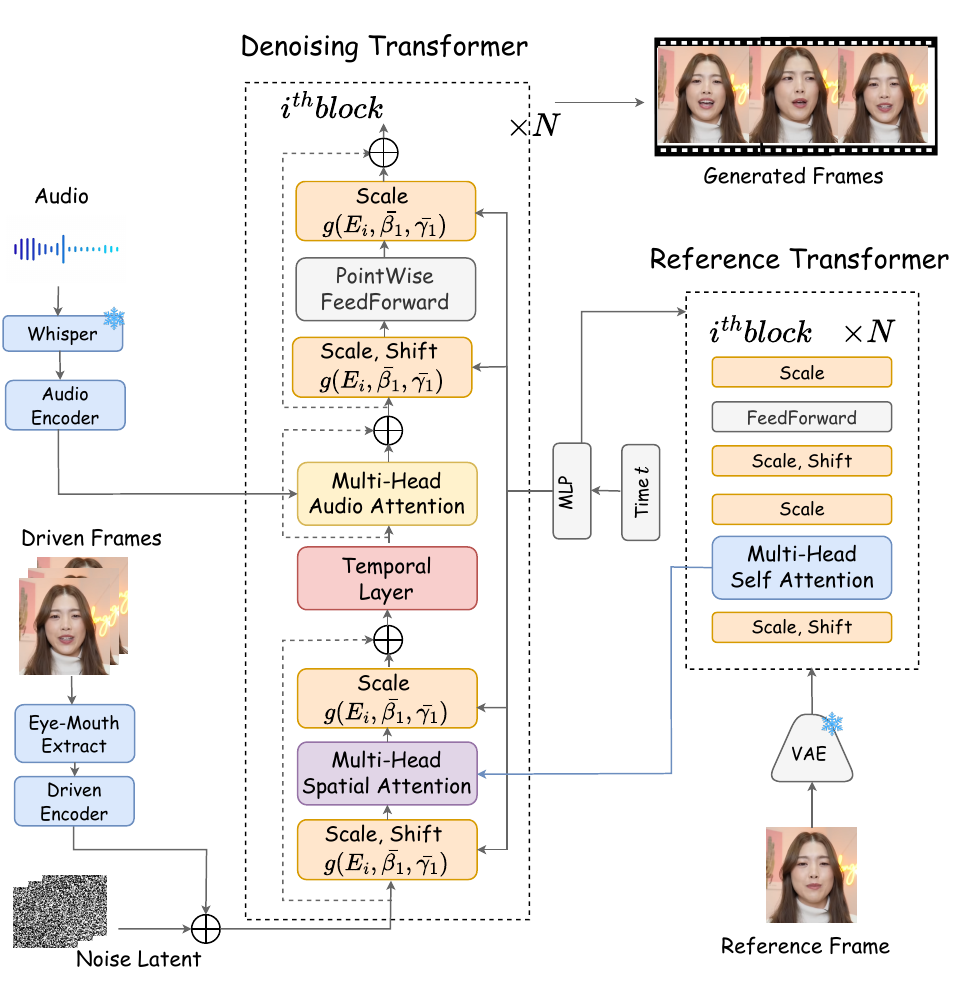}
    \caption{Mixed-modal DiT architecture of MegActor-$\Sigma$.}
    \label{fig:framework}
\end{figure}

The overall framework is summarized in Figure \ref{fig:framework}.
Specifically, MegActor-$\Sigma$ includes the Denoising Transformer and the Reference Transformer.
The Denoising Transformer is designed based on \textsc{PixArt-$\alpha$}~\cite{chen2023pixart}, aiming to flexibly infuse the input audio signals and the visual features extracted from the Reference Transformer.
The Reference Transformer, which is architecturally identical to the Denoising Transformer, aims to extract fine-grained identity and background details from the reference image.
For the driving video, we utilize Driven Encoder, composed of multiple layers of 2D convolutions, to extract motion features. 
Whisper~\cite{radford2023robust} is employed to extract speech audio features, which are then injected into the Denoising Transformer through cross-attention mechanisms. 
To enhance temporal coherence between generated frames, we integrate a temporal module into the Denoising Transformer and finetune it independently. 
Based on the framework, we propose a “Modality Decoupling Control” training strategy to balance the dominant visual modality signals with the weak audio modality signals. 
Besides, to modulate the control motion amplitude across different modalities, we introduce the ``Amplitude Adjustment" inference strategy.

\subsection{MegActor-$\Sigma$}

\textbf{Driven Encoder.}\label{DrivenEncoderSection}
Inspired by AnimateAnyone~\cite{hu2023animate}, we employ a lightweight Driven Encoder, consisting of four 2D convolutional layers to extract motion features from the driving video. 
These motion features are then aligned to the same resolution as the noise latents obtained from random sampling. 
We concatenate the motion features with the noise latents along the channel dimension. 
During training, we reinitialize the parameters of the conv-in layer of the Denoising Transformer, retaining the parameters of the first four channels and initializing the remaining channels to zero. 
This strategy mitigates the disruption to the spatial structure of the Denoising Transformer, originally initialized from \textsc{PixArt-$\alpha$}, caused by the introduction of motion features.

\noindent \textbf{Spatial Attention Layer.}\label{SpatialAttentionLayerSection}
We utilize the Reference Transformer to extract spatial features from the latent representation of the reference image which is yielded by VAE~\cite{kingma2013auto,rombach2022high}. 
The spatial features are then integrated into the Denoising Transformer via multiple self-attention layers, termed the Spatial Attention, following MagicAnimate~\cite{xu2023magicanimate}.

\noindent \textbf{Audio Attention Layer.}\label{AudioAttentionLayerSection}
The Denoising Transformer is
designed based on \textsc{PixArt-$\alpha$}, where each block originally contains a self-attention layer and a cross-attention layer. 
We replaced the original cross-attention layer with a modified version, termed the audio attention layer, which features a different hidden dimension. The audio attention layer processes audio features encoded by Whisper~\cite{radford2023robust}.

\noindent \textbf{Temporal Layer.}\label{TemporalLayerSection}
Researches~\cite{, guo2023animatediff, xu2024easyanimate} have shown that incorporating additional temporal modules into text-to-image (T2I) models for video generation can effectively capture temporal dependencies between frames and enhance their continuity. This design allows for the transfer of pretrained image generation capabilities from the base T2I model. Inspired by these findings~\cite{xu2024easyanimate}, we integrate a temporal module after each self-attention layer in the Denoising Transformer to facilitate temporal fusion between frames.

\begin{figure*}
    \centering
    \includegraphics[width=0.92\linewidth]{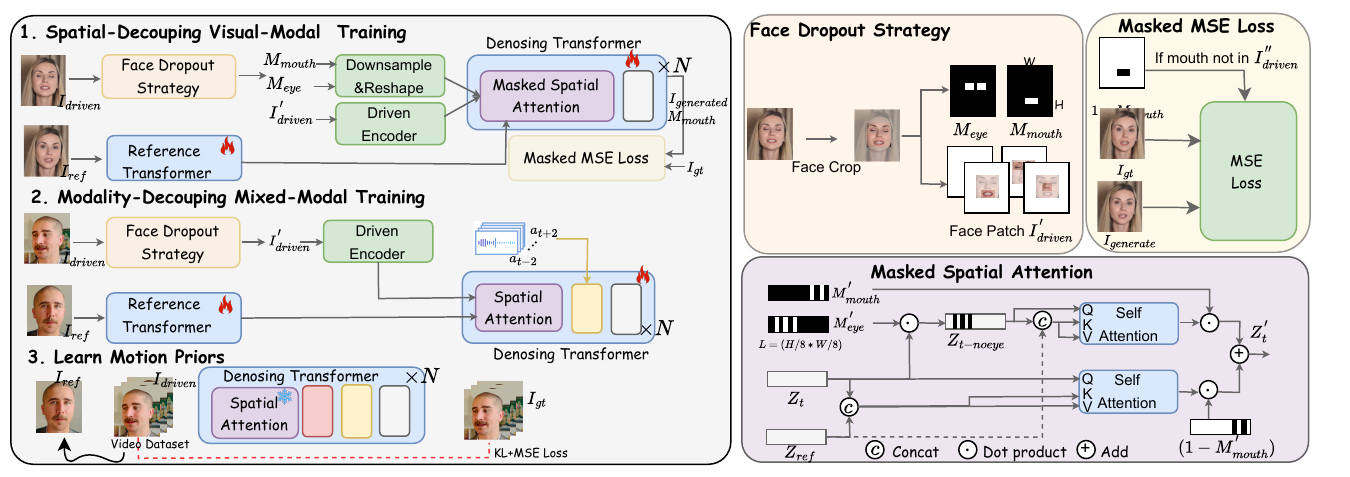}
    \caption{
    The overall framework of ``Modality Decoupling Control"  training strategy. 
    Firstly, we utilize face dropout to control partial signals (e.g., eyes or mouth). Masked spatial attention and maksed MSE loss are then applied to ensure that the control of the mouth region is decoupled.
    Secondly, we integrate audio for mixed-modal control with face dropout to dynamically balance control strength of the audio and visual modalities. 
    Finally, temporal layers are further introduced to learn motion priors. 
    }
    \label{fig:TrainingStrategy}
\end{figure*}

\subsection{Training Strategy}\label{TrainingSection}

As previously noted in Figure \ref{fig:unbalance}, training with mixed-modal control signals often leads to the neglect of audio cues when visual control is applied. We attribute this issue to visual leakage caused by the coherent nature of visual control signals, where the motion of each facial region becomes predictable based on the others. As a result, the audio signals are overshadowed, even when there is no conflicting mouth control signal in the visual modality. Therefore, the key to effective mixed-modal control lies in the spatial decoupling of the eyes control signal and mouth control signal.

To enable flexible combinations of mixed-modal control signals for MegActor-$\Sigma$, we propose the “Modality Decoupling Control” training strategy, which can be divided into three stages: ``Spatial-Decoupling Visual Training", ``Modality-Decoupling Mixed Training" and ``Motion Priors Training".

In the  ``Spatial-Decoupling Visual Training" stage, we use a face dropout strategy to generate spatial masks, randomly selecting isolated control signals, such as those from the eyes, mouth, or both. We then apply these spatial masks to control the regions where attention and the loss function are applied, ensuring that eye signals do not affect the mouth area.
In the ``Modality-Decoupling Mixed Training" stage, we further integrate audio modality and mixed-modality for combined control. Specifically, the audio modality excludes visual control signals, while the mixed modality comprises audio signals, as in the audio modality, and visual signals that focus solely on the eye region, ensuring that it does not overshadow mouth movements controlled by the audio. This approach aligns the spatial position of the mouth with the audio modality, easing the challenge of modeling mouth shapes.
In the ``Motion Priors Training" stage, we freeze the attention parameters of the visual and audio modalities and train only the motion module to achieve smooth temporal predictions. We trained our model using the exact same loss functions as \textsc{PixArt-$\alpha$}, which comprises both KL loss and MSE loss.
Detailed information about the training process can be found in Figure \ref{fig:TrainingStrategy}.

\subsubsection{Spatial-Decoupling Visual Training}
For a video containing facial movements, we randomly select two frames: one as the reference image, $I_{ref}$, with a static portrait and the other as the predicted result, termed the $I_{gt}$. 
We leverage DWPose~\cite{yang2023effective} to compute facial landmarks for the $I_{gt}$. 
During training, random region masks are selected for driving, which can be any combination of the eye and mouth patches $M_{e}, M_{m}$.
The driving frame, $I_{driven}$, will be calculated using the following formula:
\begin{equation}
\begin{aligned}
\label{I_d_Calculate}
I_{driven} = M_{driven} \cdot I_{gt}, \text{where } M_{driven} \subseteq \{M_{e}, M_{m}\}
\end{aligned}
\end{equation}
At this stage of training, the model is initialized from \textsc{PixArt-$\alpha$} and does not include the audio attention layer, thus no audio information is introduced. The training objective is to reconstruct the $I_{gt}$ based on the $I_{driven}$ and the $I_{ref}$. 
During the training process, we introduce an attention mask within the spatial attention mechanism, which is activated only when there is no mouth patch in the driving images. In this case, the embedding of the mouth region will not interact with the driving region through attention, while the rest of the regions remain consistent with the original attention~\cite{chang2023magicdance}. The mouth region attention formula is as follows:
\begin{equation}
\begin{aligned}
\label{AttentionMASK}
&Attention(Q,K,V,M_{e},M_{m}) = \neg M_{m} \times \sigma(\frac{QK^{T}}{\sqrt{d}})V \\
& + M_{m} \times \sigma(\frac{\neg M_{e}Q  (\neg M_{e}K)^{T}}{\sqrt{d}})(\neg M_{e}V)
\end{aligned}
\end{equation}
where $\sigma(\cdot)$ denotes the Softmax function, and $\neg M$ denotes the complement of the mask $M$, selecting the inverse regions not covered by the original mask. We also incorporate a spatial mask for MSE loss, which calculates the loss only for the specific regions of interest.
For example, when the driving image is eye patch, the attention mask for the mouth tokens during spatial attention excludes the eye area and the MSE loss is computed only for the regions outside of the mouth, as shown in Figure \ref{fig:TrainingStrategy}. 
This ensures that the generation of the mouth region is not controlled by the eye region.
To further mitigate information leakage arising from content overlap in specific regions between the $I_{driven}$ and $I_{gt}$, we employ face-swapping~\cite{modelscope} and stylization~\cite{podell2023sdxl} on the $I_{driven}$. Furthermore, we perform data augmentation on the driving images, including resolution adjustments, size modifications, and color transformations. 
During training, we optimize all parameters of the Denoising Transformer, Reference Transformer, and Driven-Encoder.

\subsubsection{Modality-Decoupling Mixed Training}
The model is initialized with the training parameters from the first stage, incorporating an audio attention layer into each transformer block. Using audio signals from the $I_{gt}$ as audio control conditions, the training objective is to reconstruct the $I_{gt}$ based on the $I_{driven}$, audio conditions. We train all parameters of the Denoising Transformer, Reference Transformer, and Driven-Encoder, while keeping the encoder parameters of Whisper~\cite{radford2023robust} frozen. During training, we use 10\% of visual modality data with face dropout, 20\% of audio modality data and 70\% of mixed-modality data.

\subsubsection{Motion Priors Training}
We employ the EasyAnimate-V1\cite{xu2024easyanimate} method to integrate an inter-frame attention module into the model. The training strategy remains the same as in the second stage, with the sole difference being that only the newly inserted motion module is trained.

\subsection{Inference Strategy}\label{InferencingSection}
\subsubsection{Visual control}
For visual modality control, we first derive the warp transform matrix between the reference image and each frame of the driving video given facial landmarks of all frames. 
Given the facial landmarks $KP_{0}$ of the first frame and $KP_{i}$ of the $i$-th frame, $M_{i}$ is the corresponding warp transform matrix to project $KP_{0}$ to $KP_{i}$. 
\begin{equation}
\begin{aligned}
\label{MotionScale2}
& M = \begin{pmatrix}
    a, & b, & t_{x}, \\
    c, & d, & t_{y} \\
\end{pmatrix} \\
& \theta_{x} = arctan2(b, a), \theta_{y} = arctan2(d, c)  \\
& \lambda_{x} = \sqrt{a^2 + b^2}, \lambda_{y} = \sqrt{c^2 + d^2} \\
& M' = \begin{pmatrix}
    \lambda_{x} * cos(\alpha \theta_{x}), & \lambda_{x} * sin(\alpha \theta_{x}), & \alpha  t_{x}, \\
    \lambda_{y} * cos(\alpha \theta_{y}), & \lambda_{y} * sin(\alpha \theta_{y}), & \alpha  t_{y} \\
\end{pmatrix}
\end{aligned}
\end{equation}
As depicted in Eq. \ref{MotionScale2}, we first extract rotation from the transform matrix, and scale rotation together with translation with motion control factor $\alpha$ to obtain a modified Matrix $M'_{i}$. Then we warp the $i$-th frame back using a inverse of $M_{i}$ and obtained $I'_{i}$. We add this extra step to best preserve the look of the original frame. Finally we apply $M'_{i}$ to $I'_{i}$ and get the motion-scaled frame for inference. This strategy allows for precise control over the image modality at varying levels of intensity. 
\subsubsection{Audio control}
For the audio modality, $audio\_scale$ is employed as a weighting factor during the residual summation of outputs from each Multi-head Audio Attention (MHAA), thereby adjusting the influence of audio control.
\begin{equation}
\begin{aligned}
\label{AudioScale}
& f = f + audio\_scale \times f_{MHAA}
\end{aligned}
\end{equation}
As elucidated in Eq. \ref{AudioScale}, the $f$ is the input and skip feature of the MHAA module, and $f_{MHAA}$ is the output feature of the MHAA module.

\begin{figure*}
    \centering
    \includegraphics[width=0.99\linewidth]{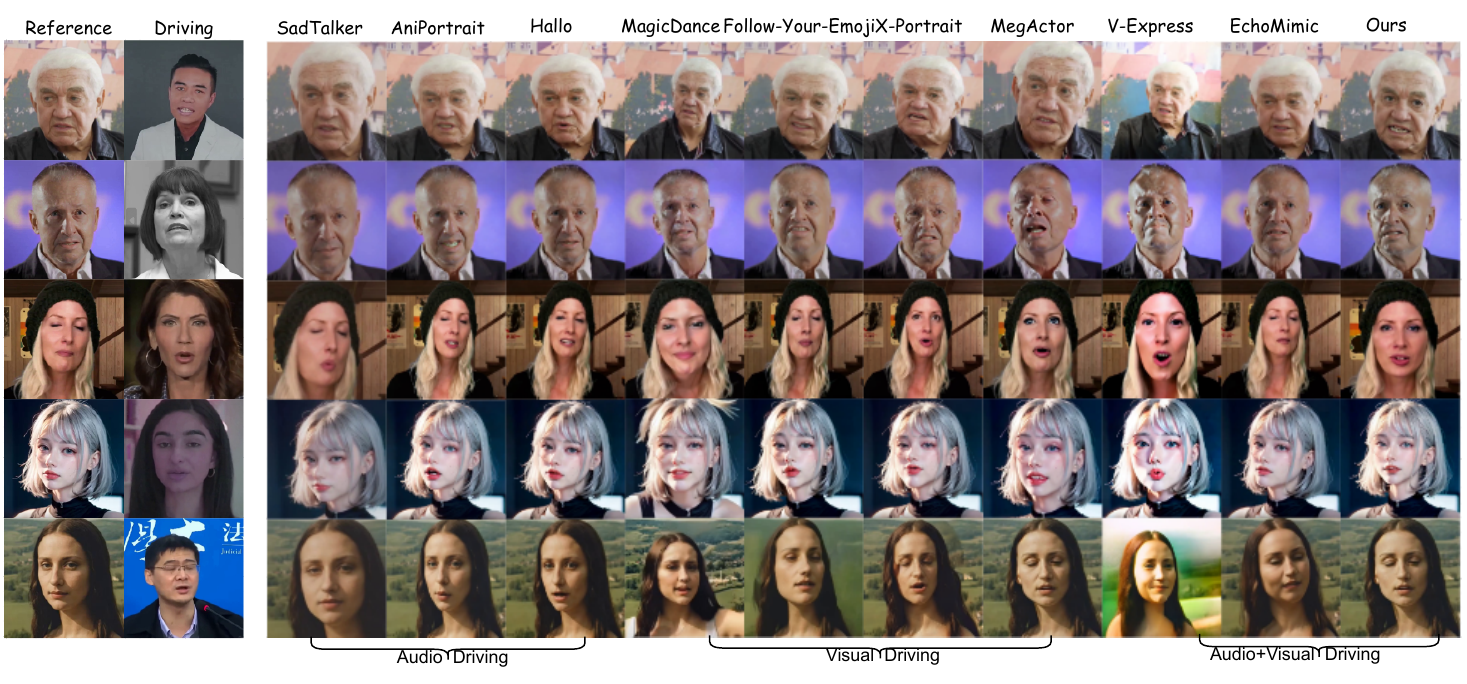}
    \caption{
    \textbf{Qualitative comparisons.}
    Proposed MegActor-$\Sigma$, driven by original images and audio, achieves accurate facial expression transfer (e.g., consistent head and eye movements) and exhibits precise identity resemblance (e.g., facial shape).
    The reference portraits in rows 1-3 are from VFHQ~\cite{xie2022vfhq}, and the rest are from the Internet.
    }
    \label{fig:compare}
\end{figure*}

\section{Public Data Process}
For diffusion models, the quality of the data significantly impacts the effectiveness of the model. However, current state-of-the-art portrait animation methods rely on their private datasets, which hinders community reproducibility and further research. 
We believe that public data still holds great potential, so we designed a series of evaluation metrics for dataset quality and used them to conduct extensive screening on large-scale multi-modal portrait animation datasets. 
Ultimately, the models trained on our filtered, high-quality public dataset outperformed those trained on private data.
Specifically, we utilized five large-scale multimodal public datasets~\cite{nagrani2017voxceleb, chung2018voxceleb2, sung2024multitalk, wang2021facevid2vid, porgali2023casual, hazirbas2021casual}, which collectively contained 4,670 hours of raw video footage. After data screening, we retained 313 hours of footage.
The evaluation metrics, including effective facial resolution, lip-to-audio synchronization, and head rotate angle, are designed as follows.

\begin{table}
\centering
\resizebox{0.49\textwidth}{!}{
\begin{tabular}{c|c|c|c|cc|c}
\hline
Datasets & \#Hours & \#Filter-Hours & Facial Res.$\uparrow$ & Sync-C$\uparrow$ & Sync-D$\downarrow$ & Head Rotation Ang.$\downarrow$ \\ \hline
VoxCeleb & 2794 & 140 & 400 $\times$ 400 & 6.218 & 8.500 & 19.4°  \\
Talking-Head 1k & 1000 & 80 & 400 $\times$ 400 & 4.956 & 9.315 & 23.7° \\
MultiTalk & 420 & 34 & 300 $\times$ 300 & 6.002 & 8.687 & 20.6° \\
CCv2 & 440 & 44 & 600$\times$ 600 & 4.345 & 9.999 & 8.2° \\
HDTF & 15.8 & 15 & 600 $\times$ 700 & 7.957 & 7.330 & 5.2° 
\\ \hline
\end{tabular}
}
\caption{
Quantitative comparison of dataset quality.
``Facial Res." indicates the average resolution of the facial region, ``Sync-C" and ``Sync-D" are used to describe the lip-audio synchronization accuracy.
``Head Rotation Ang." indicates the average head rotation angle of the video clips.
}
\label{tab:data_fliter}
\end{table}

\begin{table*}[t]
\centering
\resizebox{0.99\linewidth}{!}{
\begin{tabular}{ccc ccc ccc ccc ccc}
\toprule[1.5pt]
\multirow{2}[2]{*}{Method} & \multirow{2}[2]{*}{Framework} & \multirow{2}[2]{*}{Params} & \multirow{2}[2]{*}{Private Data} & \multirow{2}[2]{*}{Modal} & \multicolumn{5}{c}{\textbf{HDTF}} & \multicolumn{5}{c}{\textbf{CCv2}} \\
\cmidrule[0.5pt](lr){6-10} \cmidrule[0.5pt](lr){11-15}
& & & &
& FID$\downarrow$ & FVD$\downarrow$ & LPIPS$\downarrow$ & Sync-C$\uparrow$ & Sync-D$\downarrow$
& FID$\downarrow$ & FVD$\downarrow$ & LPIPS$\downarrow$ & Sync-C$\uparrow$ & Sync-D$\downarrow$
\\
\midrule[1pt]
AniPortrait & \multirow{2}[2]{*}{SD1.5} & 2.5B & No & \multirow{2}[2]{*}{A} & 33.317& 576.48  & 0.1651 & 4.0886 & 10.316 & 37.437 & 990.78 & 0.2476 & 3.6917 & 10.557 \\
Hallo &   & 2.4B & Yes & & 35.649 & 839.22 & 0.1147 & 6.7103 & 8.1199 & 38.236 & 988.29 & 0.1558 & 6.0820 & 8.8416 \\
\midrule[1pt]
MagicDance  & \multirow{4}[2]{*}{SD1.5} & 4.1B & No & \multirow{4}[2]{*}{V} & 38.399 & 707.25 & 0.1861 & 1.4071 & 12.833 & 37.893 & 794.03 & 0.2016 & 1.1953 & 12.728 \\
MegActor  &  & 2.1B & No & & 32.984 & 320.42 & 0.0889 & 5.8452 & 8.8823 & 36.979 & 402.27 & 0.1471 & 5.5078 & 9.2506 \\
Follow-Your-Emoji &  & 2.2B & Yes & & 35.063 & 365.84 & 0.1090 & 5.8462 & 8.5805 & 36.726 & 420.90 & 0.1839 & 4.8391 & 9.5500 \\
X-Portrait  &  & 2.9B  & Yes & & 32.570 & 343.52 & 0.0885 & 5.4879 & 9.1072 & 37.318 & 491.82 & 0.1588 & 5.1656 & 9.5267 \\
\midrule[1pt]
V-Express & \multirow{2}[2]{*}{SD1.5} & 2.2B & Yes & \multirow{2}[2]{*}{A+V} & 33.770 & 376.38 & 0.0960 & 6.0445 & 8.4650 & 36.746 & 405.98 & 0.1336 & 5.6041 & 9.2320 \\
EchoMimic & & 2.1B & Yes & & 31.822 & 317.04 & 0.0873 & 6.2166 & 8.3247 & 35.943 & 388.11 & 0.1280 & 6.1575 & 8.9246 \\
\midrule[1pt]
\textbf{Ours} &  DiT & \textbf{1.4B} & No & A+V & \textbf{31.497} & \textbf{302.87} & \textbf{0.0812} & \textbf{6.8226} & \textbf{8.1026} & \textbf{35.118} & \textbf{345.66} & \textbf{0.1030} & \textbf{6.5407} & \textbf{8.6185} \\
\bottomrule[1.5pt]
\end{tabular}
}
\caption{
Quantitative comparison of our MegActor-$\Sigma$ with SOTA portrait animation methods on the HDTF and CCv2 Benchmark. 
Our proposed method achieves superior results with only public dataset training, surpassing those methods trained on private datasets.
``A" denotes audio control. ``V" denotes visual control.
``A+V" denotes the combined of ``A" and ``V". 
}
\label{tab:quantitative_SOTA}
\end{table*}

\noindent \textbf{Effective Facial Resolution:} 
The quality of the generated dataset is highly dependent on high-resolution facial video dataset. if the face occupies only a small portion of the frame in high-resolution videos, the resolution of the cropped facial images remains low. 
Therefore, we employed DWPose~\cite{yang2023effective} to compute the facial bounding box and its size, with a requirement that the resolution of the detected faces must be greater than 600 $\times$ 600.

\noindent \textbf{Lip Sync Accuracy:} 
For audio control, precise alignment between audio and lip movement directly impacts the control of mouth region movements by the audio. Therefore, we use SyncNet~\cite{Chung16a} to compute Sync-C and Sync-D, which serve as measures of audio-lip synchronization accuracy. Higher Sync-C values and lower Sync-D values indicate better synchronization. We require that the Sync-C value for the video be greater than 6, and the Sync-D value be less than 8.5. 

\noindent \textbf{Head Rotation Angle:} 
The proportion of head rotation angles directly affects the difficulty of training portrait animation methods. While larger head rotation angles can help diversify the dataset, a higher proportion of such data significantly increases the training complexity. Therefore, we calculate the head rotation angle for a video clip based on landmarks, and if the average angle exceeds 30 degrees, the video is excluded.

We calculated the average performance of the datasets across these metrics, which highlights the focus and quality differences among various datasets. Ultimately, we retained 6.7\% of the public data for training, as illustrated in Tab. \ref{tab:data_fliter}.

\section{Experiment Result}
\subsection{Implementation details}
The experiments were conducted on 8 NVIDIA V100 GPUs, encompassing both the training and inference phases. Each of the three training stages consisted of 30,000 steps, with the video resolution set to 512 $\times$ 512 and a learning rate of 1e-5. Denoising Transformer is composed of 28 basic transformer blocks. To reduce computational complexity and the number of parameters, we inject spatial features from the Reference Transformer into the Denoising Transformer only in the last sixteen blocks. Among the basic transformer blocks numbered 0 through 27, we insert a temporal module only in those with odd-numbered indices. 

\subsection{Experimental Setup}
\noindent \textbf{Evaluation Metrics:} The evaluation metrics for portrait animation methods include Fréchet Inception Distance (FID), Fréchet Video Distance (FVD), Learned Perceptual Image Patch Similarity (LPIPS), Synchronization-C (Sync-C) and Synchronization-D (Sync-D). FID, FVD and LPIPS measure the similarity between generated images and real data, with lower values indicating superior performance and more realistic outputs.
Sync-C and Sync-D evaluate lip synchronization in generated videos, assessing synchronization from both content and dynamic perspectives. 

\noindent \textbf{Baselines:} In quantitative experiments, we compared our method with the publicly available implementation of audio-driven methods (Hallo~\cite{xu2024hallo}, AniPortrait~\cite{wei2024aniportrait}, SadTalker~\cite{zhang2023sadtalker}), image-driven methods (X-Portrait~\cite{xie2024x}, Follow Your Emoji~\cite{ma2024follow}, MagicDance~\cite{chang2023magicdance}), and multi-modal-driven methods (V-Express\cite{wang2024v}, EchoMimic~\cite{chen2024echomimic}). 
Evaluations were carried out on HDTF~\cite{zhang2021flow} dataset and CCv2~\cite{porgali2023casual} dataset. To ensure a rigorous evaluation, the identity data was partitioned following the standard $9:1$ ratio, with $90\%$ allocated to the training phase.

\subsection{Evaluations and Comparisons}
Table~\ref{tab:quantitative_SOTA}  presents comprehensive quantitative evaluations of various portrait animation methods on the HDTF and CCv2 benchmark. Our proposed method demonstrates outstanding performance across multiple metrics, with training on public datasets, outperforming methods trained on private data. 
Figure \ref{fig:compare} illustrates that audio-driven methods struggle to achieve consistency in head movements, eye gaze, etc. Methods driven by landmarks in visual modality and multi-modal approaches (such as MagicDance, Follow-your-emoji, V-Express, and EchoMimic) encounter issues with identity resemblance, as demonstrated in the last column Mona Lisa generation results.
Proposed MegActor-$\Sigma$ combines the accuracy and fidelity of the original images control in the visual modality with the natural smoothness of the audio modality, achieving the most superior results.

\subsection{Ablation Studies}
\textbf{Multi-Modal Control Signals:} 
Table~\ref{tab:modal_signal} presents the results of uni-modal and multi-modal control tested on the HDTF dataset. The findings indicate that the audio control method (A) imposes the weakest constraints, leading to greater discrepancies from the original video, yet it still achieves high lip synchronization. The method using both audio and image frames (A+V) is constrained by the original image frames and audio, resulting in a generated video that is most similar to the original and attains the highest lip synchronization, thus delivering the best outcomes.

\begin{table}
\centering
\resizebox{0.45\textwidth}{!}{
\begin{tabular}{c|cc|c|cc}
\hline
Modal & FID$\downarrow$ & FVD$\downarrow$ & LPIPS$\downarrow$ & Sync-C$\uparrow$ & Sync-D$\downarrow$ \\ \hline
A & 34.233 & 383.87 & 0.1168 & 6.0315 & 8.9535                   
\\ 
V & 32.838 & 313.33 & 0.0881 & 6.4513 & 8.4818
\\ 
A+V & \textbf{31.497} & \textbf{302.87} & \textbf{0.0812} & \textbf{6.8226} & \textbf{8.1026}
\\ \hline
\end{tabular}}
\caption{
Quantitative ablation comparison of modalities on HDTF benchamrk. ``A" denotes audio control. ``V" denotes visual control.
``A+V" denotes the combined of ``A" and ``V".
}
\label{tab:modal_signal}
\end{table}

\section{Conclusion}

In this paper we have presented MegActor-$\Sigma$: a mixed-modal conditional diffusion transformer (DiT) designed to unlock the full potential of versatile mixed-modality control in portrait animation. 
By addressing the challenges associated with balancing the control strengths of the audio and visual modalities, our method introduces ``Modality Decoupling Control" training strategy and ``Amplitude Adjustment" inference strategy, enabling more flexible and nuanced control.
To further facilitate extensive studies in this field, we design several dataset evaluation metrics to filter out public datasets and solely use this filtered dataset to train MegActor-$\Sigma$.
Extensive experiments demonstrate the superiority of our approach in generating vivid portrait animations, outperforming previous closed-source methods.
We hope this work will inspire the open-source community.
%

\bibliography{main}

\end{document}



\twocolumn[{%
\renewcommand\twocolumn[1][]{#1}%
\begin{center}
    \centering
    \includegraphics[width=0.99\linewidth]{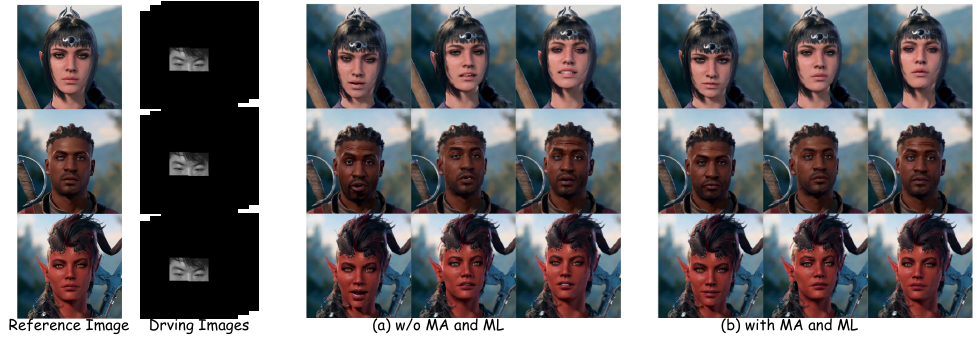}
    \captionof{figure}{
    The ablation study on masked attention (MA) and masked MSE loss (ML). As shown in (a), when Spatial-Decouping Visual-Modal training strategy with MA and ML is not employed, the generated results exhibit a mouth movement pattern, even in the absence of an mouth control signal. However, after employing proposed MA and ML, as shown in (b), the generated results no longer display any mouth movement. 
    This demonstrates the effectiveness of our training strategy.
    }
\label{fig:ablation}
\vspace{2mm}
\end{center}%
}]

\section{More Details on Ablation Study}
The table~\ref{tab:Method_Ablate} presents an evaluation of the contribution of each proposed component. During training, we replaced Masked Attention with the original spatial self-attention mechanism from DiT, denoted as w/o MA. Additionally, Mask Loss for MSE was excluded, denoted as w/o ML. We also tested the exclusion of both Masked Attention and Masked Loss for MSE, denoted as w/o MA and ML. The impact of data filtering operations on model performance was assessed by training the model with unfiltered data, denoted as w/o Filter. We further explored different model architectures, comparing the performance of the DiT architecture and the SD1.5 architecture under our training regimen and data. When the SD1.5 architecture was used, denoted as w/o DiT, Mask Attention was applied only to the transformer blocks within SD1.5, and the loss function was replaced with Mask Loss applied to MSE. Ablation experiments were performed on 
 HDTF benchmark. To more clearly illustrate the influence of individual components on mixed-modal portrait animation performance, all tests were conducted using audio and the facial eye region in visual modality.

As demonstrated in Tab. \ref{tab:Method_Ablate}, the experimental results demonstrate the following: 1) When spatial decoupling of the visual modality is not applied during training, audio control becomes dominant, and the outcomes approach the performance level of single-visual modality training, as evidenced by the comparison between ``V" unimodal (Tab. 4) and w/o MA and ML (Tab. \ref{tab:Method_Ablate}) with respect to FID scores (32.838 vs 32.253) and LPIPS (0.0881 vs 0.0876).
2) Without the filtering of public data, the model's performance significantly declines due to the presence of a large amount of low-quality data, as reflected in the FID score (34.967 vs \textbf{31.497}) and the Sync-C score (4.1260 vs \textbf{6.8226}).
3) Compared to SD1.5, DiT shows an improvement in metrics, enabling the generation of higher quality images, which is corroborated by the FID score (32.263 vs \textbf{31.497}). 
This validates the effectiveness of our proposed modules.

We further visualize the effectiveness of our proposed Spatial-Decouping Visual-Modal training strategy in addressing the issue where the mouth area is controlled by the eye's control signal. As shown in Fig. \ref{fig:ablation}, without using our proposed training strategy with masked attention and masked loss, the generated mouth moves in a regular pattern even in the absence of a mouth control signal. After training with our proposed strategy, in the absence of a mouth control signal, the mouth remains stationary, achieving spatial decoupling of the controlled regions. This provides a prerequisite for subsequent mixed control of audio and visual modalities, effectively avoiding the dominance of the visual modality over the audio modality,  and achieving flexible and vivid mixed-modal portrait animation.

\begin{table}
\vspace{6mm}
\centering
\resizebox{0.49\textwidth}{!}{
\begin{tabular}{c|cc|c|cc}
\hline
Methods & FID$\downarrow$ & FVD$\downarrow$ & LPIPS$\downarrow$ & Sync-C$\uparrow$ & Sync-D$\downarrow$ \\ \hline
w/o MA & 32.438 & 317.36 & 0.0892 & 6.4319 & 8.4976                  
\\ 
w/o ML & 32.215 & 314.17 & 0.0863 & 6.4227 & 8.5059
\\ 
w/o MA and ML & 32.253 & 310.97 & 0.0876 & 6.3997 & 8.5235
\\ 
w/o Fliter & 34.967 & 791.01 & 0.1436 & 4.1260 & 10.426                   
\\
w/o DiT & 32.263 & 345.34 & 0.0868 & 6.6765 & 8.3378
\\ \hline
MegActor-$\Sigma$ & \textbf{31.497} & \textbf{302.87} & \textbf{0.0812} & \textbf{6.8226} & \textbf{8.1026}
\\ \hline
\end{tabular}}
\caption{
Quantitative ablation study on different components on HDTF benchmark.
The results show that omitting spatially decoupled visual modality training leads to audio control dominating, resulting in performance closer to a single visual modality. 
Skipping the filtering process on public data introduces a significant amount of low-quality data, causing a notable drop in model performance. 
DiT outperforms SD1.5 across metrics, particularly by achieving lower FID and FVD scores, which reflect higher-quality generated images.
``MA" denotes masked attention. ``ML" denotes masked MSE loss.
``Filter" denotes data filtering process. 
}
\label{tab:Method_Ablate}
\end{table}


